\def\year{2021}\relax
\documentclass[letterpaper]{article} 
\usepackage{aaai21}  
\usepackage{times}  
\usepackage{helvet} 
\usepackage{courier}  
\usepackage[hyphens]{url}  
\usepackage{graphicx} 
\usepackage{natbib}  
\usepackage{caption} 
\usepackage{multirow}
\usepackage{amsmath}
\usepackage{mathtools, nccmath}
\usepackage{siunitx}
\usepackage{subcaption}
\nocopyright
\title{Introducing an Abusive Language Classification Framework \\for Telegram to Investigate the German Hater Community}
\author {
    Maximilian Wich\textsuperscript{\rm 1},
    Adrian Gorniak\textsuperscript{\rm 1},
    Tobias Eder\textsuperscript{\rm 1}, \\
    Daniel Bartmann\textsuperscript{\rm 1},
    Burak Enes \c{C}akici\textsuperscript{\rm 1},
    Georg Groh\textsuperscript{\rm 1}
     \\
}
\affiliations {
    \textsuperscript{\rm 1} Technical University of Munich, Munich, Germany  \\
    maximilian.wich@tum.de, adrian.gorniak@tum.de, tobias.eder@in.tum.de, \\daniel.bartmann@in.tum.de, burak-enes.cakc@tum.de, grohg@in.tum.de
}

\begin{document}

\maketitle

\begin{abstract}
Because  traditional social media platforms continue to ban actors spreading hate speech or other forms of abusive languages (a process known as deplatforming), these actors migrate to alternative platforms that do not moderate users’ content. One popular platform relevant for the German hater community is Telegram for which limited research efforts have been made so far.

This study aimed to develop a broad framework comprising (i) an abusive language classification model for German Telegram messages and (ii) a classification model for the hatefulness of Telegram channels. For the first part, we use existing abusive language datasets containing posts from other platforms to develop our classification models. For the channel classification model, we develop a method that combines channel-specific content information collected from a topic model with a social graph to predict the hatefulness of channels. Furthermore, we complement these two approaches for hate speech detection with insightful results on the evolution of the hater community on Telegram in Germany. We also propose methods for conducting scalable network analyses for social media platforms to the hate speech research community. As an additional output of this study, we provide an annotated abusive language dataset containing 1,149 annotated Telegram messages.

\end{abstract}

\section{Introduction}

Hate speech and other forms of abusive language are a severe challenge that social media platforms, such as Facebook, Twitter, and YouTube, are facing nowadays \cite{duggan2017online}. Moreover, this problem is not limited to the online world only; studies have shown that online hate correlates with physical crimes in the real world \citep{muller2021fanning,williams2020hate}, making the phenomenon a societal challenge for everybody.  

To enforce a fast reaction to harmful content on social media platforms, Germany has passed a set of laws (Network Enforcement Act) to force social media companies to take action against hate speech on their platforms  \cite{rafael2019,echikson2018germany}. These actions range from deleting single posts that contain hateful content to banning actors from the platform, which is called deplatforming \citep{hatenotfound}. While deplatforming helps limit the reach of these hate actors \citep{hatenotfound}, it often makes them migrate to less or un-regulated platforms and continue their hateful communication  \citep{roger-deplatforming,hatenotfound,urman-2020}; one such alternative social media platform is Telegram \citep{roger-deplatforming,hatenotfound,urman-2020}. In Germany, Telegram have become the focal point for right-wing extremists, conspiracy theorists, and COVID-19 deniers   \citep{hatenotfound,urman-2020,Eckert2021}. Along with this rapid increase in popularity and usage by various user types, two important challenges regarding abusive language detection arise: first, the automatic detection of abusive content in such texts and, second, an aggregated view on the account level to identify hateful accounts. For both challenges, we propose a machine learning-based approach.

Previously,  most research efforts on detecting hate speech, especially in German texts, focused on posts and comments from Twitter and Facebook  \citep{ross2016hatespeech,bretschneider2017detecting,struss2019overview,wiegand2018overview,hasoc2019,hasoc2020,wich-covid-19,wich2021} but not on Telegram. We want to bridge this gap and build abusive language classification models for Telegram messages. Because there is no abusive language dataset available that contains labeled Telegram messages in German, our approach is to use existing abusive language datasets in German collected from other platforms and construct a classification model for Telegram. This leads to the first research question for this study:

\begin{description}
\item [RQ1] Can existing abusive language datasets from other platforms be used to develop an abusive language classification model for Telegram messages?
\end{description}

Because the development of an abusive language classification model requires significant amounts of data, we collected such data from the platform (Telegram) over a longer period. By collecting these data, we could also formulate additional questions about the type of content and its spread on the social media platform. Because there is little research on these types of communication channels and their content, we were also interested in how this content has changed over a longer period, while deplatforming was occurring on other social media. Thus, we formulate an additional research question in terms of message contents:

\begin{description}
\item [RQ2] How did the prevalence of abusive content evolve in the last years on Telegram?
\end{description}

Moving away from the message-level approach and toward a user-based approach for abusive language detection, so far no methodology has been introduced to address this problem for Telegram. As a solution, we propose developing a graph model leveraging topical information for channels in a German hater community on Telegram to find suitable representations, leading to the third research question: 

\begin{description}
\item [RQ3] Can a classification model be used to predict whether a Telegram channel is hateful or not?
\end{description}

Lastly, maintaining the channel perspective, we were interested to investigate whether our approach would allow for the derivation of channel clusters and communities, which is another important aspect regarding online hate. For this, we analyzed the topical distribution and the graph embeddings for each channel, resulting in research question four:

\begin{description}
\item [RQ4] Can we leverage the topical distribution and graph embeddings to derive meaningful clusters from channels?
\end{description}
As an additional contribution, we release an abusive language dataset containing 1,149 Telegram messages labeled as \textit{abusive} or \textit{neutral}.


\section{Related Work}

Studies on Telegram are limited, but the number began to grow in the past years. \citet{baumgartner2020pushshift} released an unlabeled dataset containing 317,224,715 Telegram messages from 27,801 channels, which were posted between 2015 and 2019. They used a snowball sampling strategy to discover channels and collect messages, starting with approximately 250 seed channels (mainly right-wing channels or channels about cryptocurrency). \citet{roger-deplatforming} conducted an empirical study on actors who were deplatformed on traditional social media and migrated to Telegram. As part of their study, they used a classification model based on hatebase.org to detect messages with hateful language \cite{roger-deplatforming}. \citet{urman-2020} conducted an in-depth network analysis of a far-right community on Telegram. They used a snowball sampling strategy to uncover this community, starting with a German-speaking far-right actor. \citet{hatenotfound} analyzed German hate actors across various social media platforms and investigated the impact of deplatforming activities on these actors. According to them, "Telegram has become the most important online platform for hate actors in Germany" \cite[p. 5]{hatenotfound}. With a focus on COVID-19, \citet{Hohlfeld2021} and \citet{Holzer2021} investigated public German-speaking channels on Telegram. The only labeled abusive language dataset with Telegram messages that we found is provided by \citet{solopova2021telegram}. They released a dataset containing 26,431 messages in English from a channel supporting Donald Trump. To the best of our knowledge, no study has developed an abusive language classification model for German Telegram messages or channels. 

Because there is no annotated German Telegram dataset available, we decided to train our classification model on existing German abusive language datasets. In total, we found eight of such datasets \cite{ross2016hatespeech,bretschneider2017detecting,wiegand2018overview,struss2019overview,hasoc2019,hasoc2020,wich2021,wich-covid-19}. We decided to use five of them---which constitute the most recent ones, excluding \citet{wich2021}. These five datasets have comparable label schemata, and a large portion of the data is from the same period as our collected Telegram data. \citet{wich2021} was excluded because their data were only pseudo-labeled. More details on the selected datasets can be found in the following section.

\section{Methodology}

In the first part, we describe how we collected data from Telegram. After that, we provide details on how we developed the abusive classification model for Telegram messages based on datasets from other platforms. In the third part, we describe how we developed a classification model to predict whether a channel is a hater based on the results from the message classifier and the social graph.

\subsection{Collecting Data}
We used a snowball sampling strategy to collect data from Telegram. We only collected messages from public channels that were accessible via the website t.me. A channel is comparable to a news feed: the channel operator can broadcast messages to subscribers of the channel, but subscribers cannot directly post messages on the channel. Groups and private chats were excluded from the data collection process. As seeds for the snowball sampling strategy, we used a list of German hate actors proposed by \citet{hatenotfound}. 
At the time of data collection, 51 channels from \citet{hatenotfound}'s list were still accessible. The list comprises, among others, far-right actors, supporters of Qanon, and alternative media.

In the first round of snowball sampling, we collected messages from all seed channels. In the next round, we collected all channels that were mentioned in messages collected from the first round or whose messages were forwarded by the channels of the first round. We repeated this procedure in the third round, but we excluded all newly discovered channels due to a large number of channels. We defined a threshold: a channel must be mentioned or forwarded by at least five channels for us to collect its messages. From all channels, we collected messages that were posted between 01/01/2019 and 03/15/2021.

After data collection, we conducted language detection on the messages because the crawling process also collects other language channels such as Russian and English and we wanted to keep the focus on German. We used multilingual word vectors from \textit{fastText} to classify languages \citep{grave2018learning}. The language detection here is based on the message text and a link preview if it exists. In a second step, the language labels of messages are aggregated on a channel level. The language of a channel is German if it is the most or second most common language in the channel. The reason for the latter is that some German channels primarily share content from foreign-language sources.

\subsection{Building Classification Models for Telegram Messages}

\paragraph{Models} To classify Telegram messages, we trained several binary classification models on different German datasets. The goal is to combine multiple classifiers to improve classification performance because each dataset covers different aspects and topics of abusive languages. The reason for focusing on binary classification was that it makes combining classifiers easier.

All classification models are based on pretrained BERT base models \citep{devlin-etal-2019-bert}.  We used \texttt{deepset/gbert-base} \citep{chan-etal-2020-germans} and \texttt{dbmdz/bert-base-german-cased}\footnote{\url{https://huggingface.co/dbmdz/bert-base-german-cased}} depending on the model's performance on the individual dataset. Our hyperparameters for training the models comprise a maximum number of eight epochs, a learning rate of $5\times 10^{-5}$, and a batch size of eight. In addition, we implemented an early stopping callback that stops the training after four consecutive epochs without any improvement. We selected the model with the highest macro F1 score on the validation set.

Before training the models, texts are preprocessed. The preprocessing steps comprise, among others, masking URLs and user names and replacing emojis.

\paragraph{Data} 
We used the following German abusive language datasets collected from different platforms (mainly Twitter) to train our models:
\begin{itemize}
\item \textit{GermEval 2018}: \citet{wiegand2018overview} released an offensive language dataset as part of the shared task GermEval Task 2018. It contains 8,541 tweets with a binary label (\textit{offense}, \textit{other}) and a fine-grained label (\textit{profanity, insult, abuse, other}). We used the train/test split proposed by the authors and used a 90/10 split for the training/validation set. 
\item \textit{GermEval 2019}: \citet{struss2019overview} published an offensive language dataset that is part of the GermEval Task 2019. It comprises 7,025 tweets that are labeled with the same labeling schema, as the previous dataset, but a further dimension was added \textit{(implicit, explicit}). The data were split in the same way as GermEval 2018.
\item \textit{HASOC 2019}: \citet{hasoc2019} released a multilingual hate speech and offensive language dataset, called "Hate Speech and Offensive Content Identification in Indo-European Languages" \citep[p. 1]{hasoc2019}, as part of a shared task. It comprises posts from Facebook and Twitter in German, English, and Hindi. The German part comprises 4,669 records with a binary label (\textit{non hate-offensive, hate and offensive}) and a fine-grained label (\textit{hate}, \textit{offensive}, \textit{profanity}). We used the train/test split proposed by the authors and used a 90/10 split for the training/validation set. 
\item \textit{HASOC 2020}: \citet{hasoc2020} published another dataset, which is comparable to the previous one. It consists of posts from YouTube and Twitter in German, English, and Hindi. The German part has a size of 3,425 records using the same labeling schema as the previous dataset. We used  the proposed train/validation/test-split of 70\%/15\%/15\%.
\item \textit{COVID-19}: \citet{wich-covid-19} released an abusive language dataset containing 4,960 German tweets that primarily focus on COVID-19. These tweets have a binary label (\textit{neutral}, \textit{abusive}). We used a train/validation/test split of 70\%/15\%/15\%.
\end{itemize}

We trained individual classification models for all datasets, except for HASOC 2019 because we could not train a model that provides an acceptable classification performance. Furthermore, we combined the GermEval and HASOC datasets and trained two additional classifiers on the two combined datasets. Combining these datasets was possible because the respective datasets use the same labeling schema. 

\paragraph{Classifying Telegram Messages}

Because a Telegram message can have up to 40,986 characters, the tokenized message may exceed the maximum sequence length of the BERT model, which is 512. To tackle this problem, we split all messages that had more than 412 words into parts with a maximum length of 412 words. When splitting a message, we made sure not to split sentences. For this purpose, we used the sentence detection method of the library spaCy \cite{spacy}. There were two reasons for setting the threshold to 412 words. First, using words instead of tokens was easier during preprocessing. Second, a word can be tokenized into multiple tokens. Therefore, we set the threshold to 412 instead of 512. Every part of the split message was individually classified. The final label of the complete message results from the highest probability for the abusive class. The reason for this approach was because an abusive text can contain nonabusive sentences but not the other way around. In addition to the six classification models, we used Google’s Perspective API\footnote{\url{https://www.perspectiveapi.com/}} to classify Telegram messages. The API returns a toxicity score between 0 and 1, representing how toxic the content of a text is. We used these classification results as a baseline to benchmark our models.

\paragraph{Evaluating Classification Models}
To evaluate the classification performance of our trained models on Telegram messages, five annotators manually annotated 1,150 of the classified Telegram messages. More information about the annotators follows below. The 1,150 messages originated from two different sampling strategies. The first strategy uses the classification results of the six trained models and the Perspective API. For each classifier, we sampled 50 messages classified as abusive and 50 classified as neutral, resulting in a total of 700. The second strategy used a topic model trained on Telegram messages (more details on the topic model can be found in the subsection Topic Model). We randomly sampled 30 messages from the 15 most prominent topics. Finally, we ensured that the annotation candidates do not contain any duplicates. As a result, we assured that the dataset has a certain degree of abusive content and that it represents the most relevant topics.

We use the labeling schema of the \textit{COVID-19} dataset proposed by \citet{svenja2021} and \citet{wich-covid-19} because it is compatible with the binary schema of the \textit{HASOC} and \textit{GermEval} datasets:

\begin{itemize}
\item \textit{ABUSIVE}: The tweet comprised "any form of insult, harassment, hate, degradation, identity attack, or the threat of violence targeting an individual or a group. " \citep[p. 36]{svenja2021}
\item \textit{NEUTRAL}: The tweet did "not fall into the \textit{ABUSIVE} class." \citep[p. 36]{svenja2021}
\end{itemize}

Data were annotated by four nonexperts and one expert, who are males and in their twenties or early thirties. The annotation process consisted of three phases. In phase 1, the expert presented and explained the annotation guidelines to the four nonexperts. Subsequently, all five annotators annotated the same 50 messages. In 18 cases, the annotators did not agree on the final label. These cases were discussed in a meeting to align the five annotators. In phase 2, the annotators annotated the remainder of the 1,150 messages. Each message was annotated by two different annotators. The annotators were allowed to skip a message if they could not decide on a label. In phase 3, messages without a consensus were annotated by three additional annotators so that a majority vote was possible. We used Krippendorff’s alpha \citep{krippendorff2004content} to measure interrater reliability. To assist in annotations, we used the text annotation tool of Kili Technology \citep{kili}.

\paragraph{Combining Classification Models}
Because the datasets and consequently the classification models cover different aspects of abusive languages, we combined the six classifiers to improve classification performance (Perspective API was not part of the combination). The labels produced by this combination were used for subsequent experiments. 

\paragraph{Analyzing Evolution of Abusive Content}
We performed two analyses to evaluate the evolution of abusive content in the German hater community on Telegram to answer RQ2. First, we compared the number of abusive messages with all messages from the collected German channels between 01/01/2019 and 02/28/2021 on a monthly level. We excluded the messages posted in March 2021 because we did not have data for the entire month. Then, we examined the relative share (prevalence) of abusive content in the messages from all German channels for the same period and granularity. In addition, we reported the prevalence of abusive content from the seed channels and the 1st-degree network of the seed channels.

\subsection{Building a Classification Model for Hatefulness of Channels}

\begin{figure*}[ht]
     \centering
     \begin{subfigure}[b]{0.24\textwidth}
         \centering
         \includegraphics[width=\textwidth]{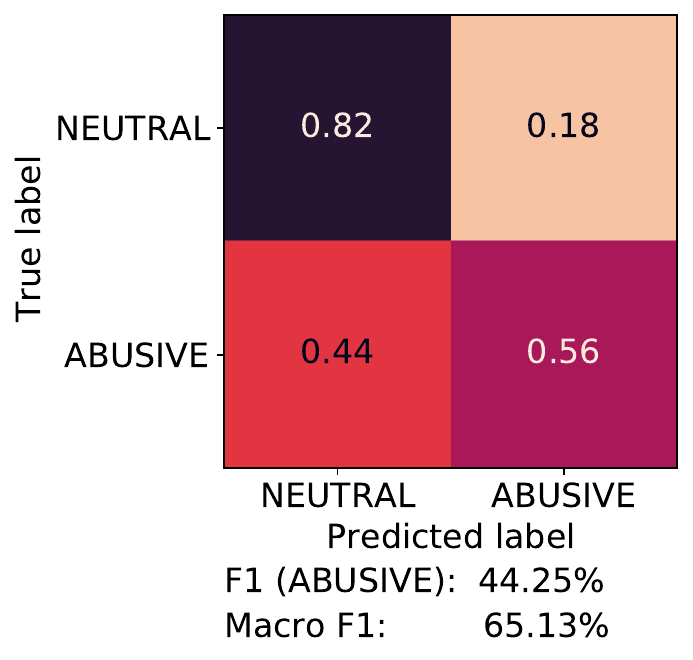}
         \caption{\textit{GermEval 2018}}
         \label{fig:ge18}
     \end{subfigure}
     \hfill
     \begin{subfigure}[b]{0.24\textwidth}
         \centering
         \includegraphics[width=\textwidth]{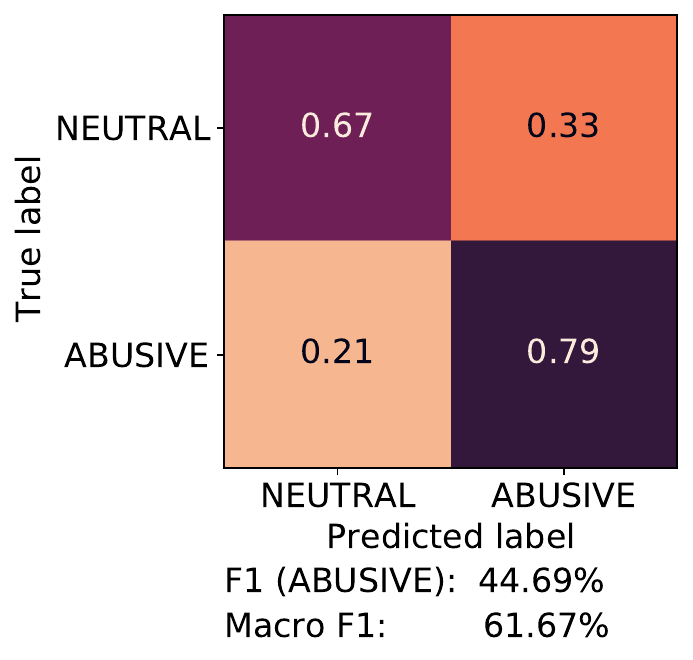}
         \caption{\textit{GermEval 2019}}
         \label{fig:ge19}
     \end{subfigure}
     \hfill
     \begin{subfigure}[b]{0.24\textwidth}
         \centering
         \includegraphics[width=\textwidth]{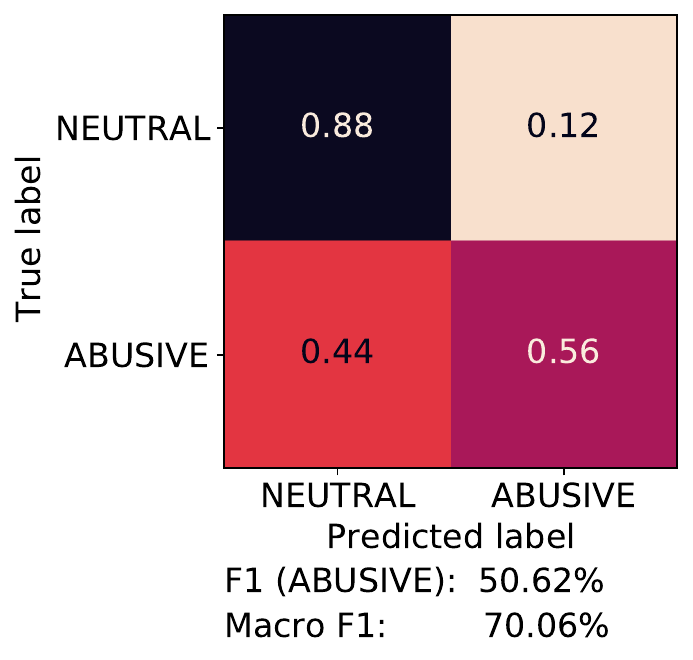}
         \caption{\textit{GermEval 18/19}}
         \label{fig:ge1819}
     \end{subfigure}
     \hfill
     \begin{subfigure}[b]{0.24\textwidth}
         \centering
         \includegraphics[width=\textwidth]{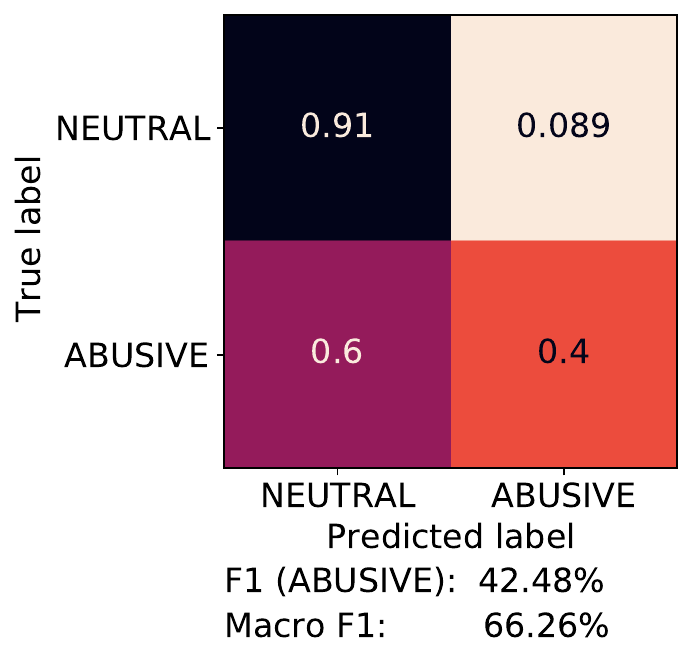}
         \caption{\textit{HASOC 2020}}
         \label{fig:ha20}
     \end{subfigure}
     \hfill
     \begin{subfigure}[b]{0.24\textwidth}
         \centering
         \includegraphics[width=\textwidth]{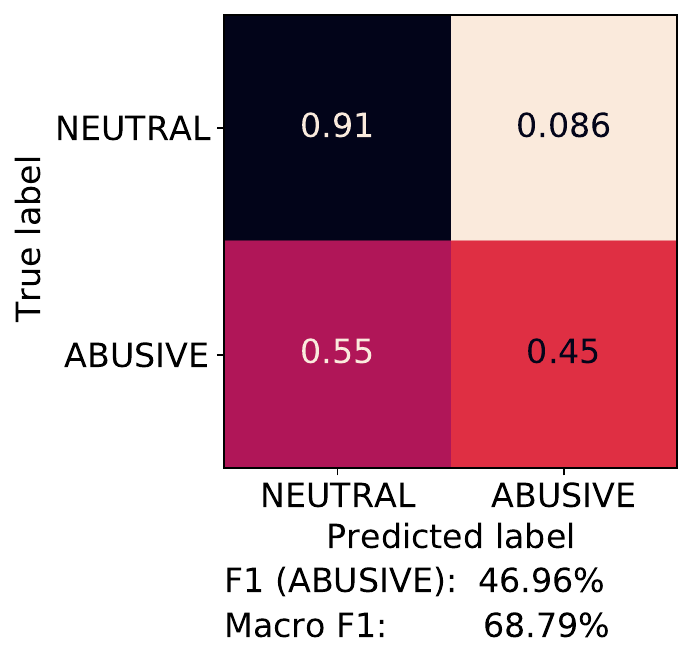}
         \caption{\textit{HASOC 19/20}}
         \label{fig:ha1920}
     \end{subfigure}
     \hfill
     \begin{subfigure}[b]{0.24\textwidth}
         \centering
         \includegraphics[width=\textwidth]{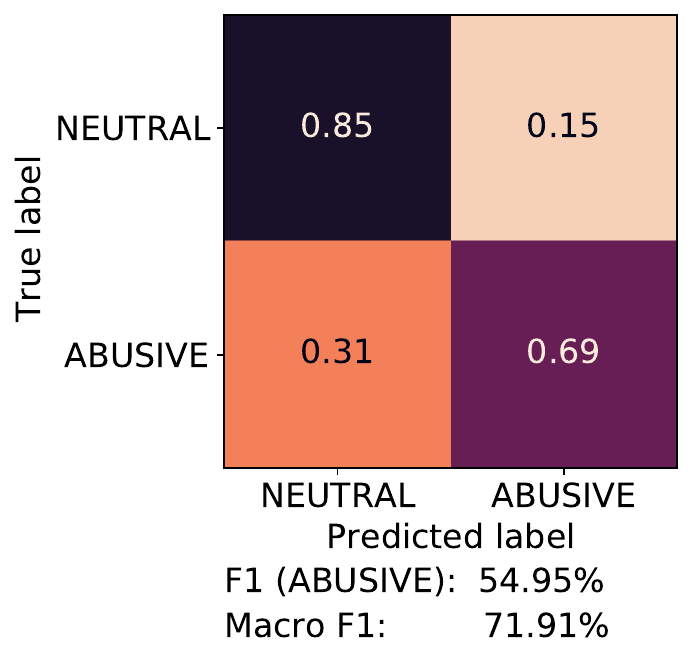}
         \caption{\textit{COVID-19}}
         \label{fig:covid-19}
     \end{subfigure}
     \hfill
     \begin{subfigure}[b]{0.24\textwidth}
         \centering
         \includegraphics[width=\textwidth]{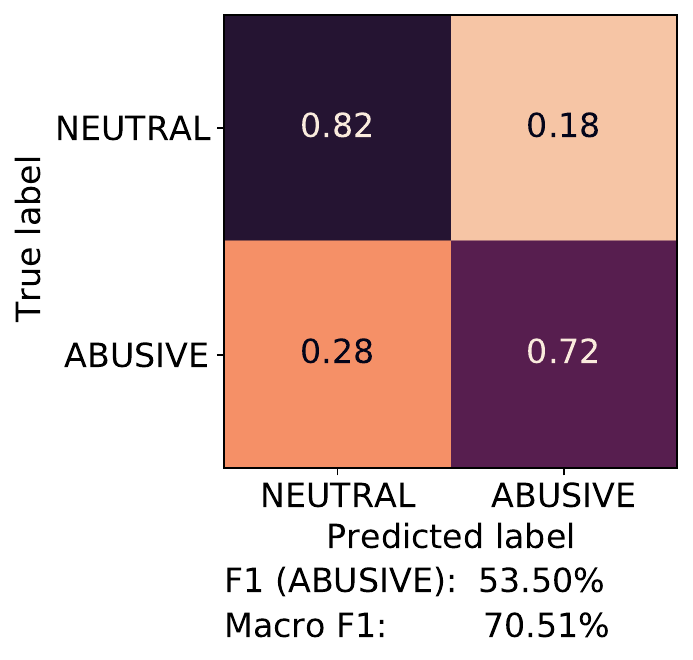}
         \caption{\textit{Perspective API (baseline)}}
         \label{fig:perspective}
     \end{subfigure}
     \hfill
     \begin{subfigure}[b]{0.24\textwidth}
         \centering
         \includegraphics[width=\textwidth]{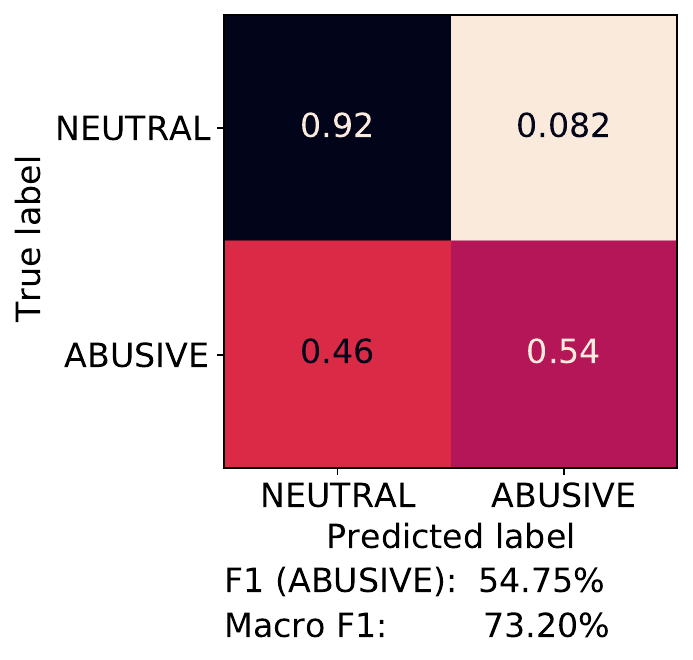}
         \caption{\textit{Best combination}}
         \label{fig:combination}
     \end{subfigure}
        \caption{Classification performance of the various models on annotated Telegram evaluation set.}
        \label{fig:matrices}
\end{figure*}

\paragraph{Channel Labels} We had to determine a label for each channel based on the abusive messages in the channel. We defined a \textit{hater} as a channel that posted or forwarded at least one abusive message. However, setting the threshold to one proved problematic due to the possibility of misclassification, meaning that false positives would cause neutral channels to be classified as haters. Instead, for each message, we calculated a threshold based on the conditional probability that a message is neutral under the condition of it being labeled as abusive. This conditional probability is retrieved from a confusion matrix (Figure \ref{fig:combination}). As a result, we had to adjust the weighting of the confusion matrix’s rows. Because we intentionally oversampled the abusive class in the evaluation set, the ratio of abusive texts was no longer representative of the entire dataset. We assume that the relative share of abusive content is 3.1\%  for 2020, based on the results from the analysis of the abusive content’s evolution. The resulting conditional probability is 82.9\%. Assuming an error rate of smaller than 5.0\% , we need at least 17 messages that are classified as  \textit{abusive} to be certain that at least one message is \textit{abusive}. Second, we created a directed graph representing the network of channels. Each channel is a node; a directed edge from nodes A to B exists if A either mentions B or forwards a message from B.

\paragraph{Topic Model} We assigned a topic distribution vector as a feature to each node, representing the topical distribution within the messages of the channel. The topical distribution was calculated on the basis of the topic model generated with Top2Vec \citep{angelov2020top2vec}. We relied on the hyperparameter selection of the author, used the \texttt{distiluse-base-multilingual-cased}\footnote{\url{https://huggingface.co/distilbert-base-multilingual-cased}} pretrained  sentence transformer as embedding model, and sampled 250,000 messages (500 messages from the 500 channels containing the largest amount of messages in our dataset) as training samples. From the 100 most relevant topics, we manually chose nine topics to serve as proxies for hateful content. They are listed in Table \ref{tab:topics}: the topic name in the first column was derived on the basis of the most descriptive terms of the respective topic vectors from which we provide the first three terms in the second column (in German). Because we are working with many channels that can be associated with German hater communities, we relied only on these topics to cluster different topical emphases with respect to potentially harmful content. We aggregated the counts of all documents in our dataset with cosine similarity to any of the selected topics greater than 0.5 and normalized these counts to create a topic distribution for each node.

\begin{table*}[ht]
\caption{Topics selected for topic distribution along with three descriptive terms of the topic model.}
\centering
\label{tab:topics}
\resizebox{0.80\textwidth}{!}{%
\centering
\begin{tabular}{lll}
\hline
\textbf{Topic} & \textbf{Descriptive terms}       & \textbf{Translation}                    \\ \hline
Vaccinations   & impfen, geimpft, durchgeimpft                & vaccinate, vaccinated, fully vaccinated \\
Police         & Polizeigewalt, Bundespolizei, Polizeif\"uhrung & police violence, federal police, police leadership \\
COVID-19       & Coronakrise, Corona, Coronaleugner   & corona crisis, corona, corona denier         \\
Migration      & Migrantengewalt, Migranten, Refugees  & migrant violence, migrants, refugees        \\
Extremism      & rechtsextremer, rechtsextremen, rechtsextreme & far-right \\
Racism         & Rassismus, rassistischer, rassistisch    & racism, racist     \\
Islamophobia   & Moslemterror, Islamisten, Islamisierung  & Muslim terror, Islamists, Islamization     \\
Violence       & sterben, Massenm\"order, Massenmord     & die, mass murder        \\
Antisemitism   & Antisemismus, Antisemiten, antisemitische  & antisemism, antisemites, antisemitic   \\ \hline
\end{tabular}%
}

\end{table*}

\paragraph{Graph Model} We used GraphSAGE to generate embeddings for the graph (\citet{hamilton2017inductive}). We used the Directed GraphSAGE method from the StellarGraph library \cite{StellarGraph}. As we were learning unsupervised embeddings, i.e., we did not provide the learning model with labels of the channels, we used the \textit{Corrupted Generator} of StellarGraph for sampling additional training data. During training, the model learned to differentiate between true graph instances and corrupted ones. The model was trained for 500 epochs with two layers of size 32 each, an Adam optimizer, and an early stop after 20 epochs of patience.

\paragraph{Channel Classification} 
We developed a neural network (NN)  classification model using the graph embeddings to predict the classes. The model consists of two densely connected NN layers. The input for the first layer is a 32-dimensional graph embedding. The second layer (output) has two units due to the binary task. The first layer uses a rectified linear unit activation function, whereas soft-max was applied to the output layer. To train the model, cross-entropy was used as a loss function with accuracy as the metric using an Adam optimizer. We trained the model for a maximum of 150 epochs with a batch size of eight with an early stopping strategy that had the patience of 100 epochs and a minimum delta of 0.05 for accuracy on the validation set. The dataset was split into training/validation/test sets (70\%/15\%/15\%). 

The dataset for RQ3 only used messages from 2020 as the social network on Telegram is rapidly evolving and changing. That means that older edges might no longer be relevant and the network structure would generally be less meaningful. Another aspect of this decision is that the emergence of COVID-19 strongly influenced and accelerated the evolution of the network, which did not exist pre-COVID-19 pandemic.

\section{Results}
\subsection{Collecting Data}

In total, we collected 13,822,605 messages from 4,962 channels that were posted after 01/01/2019 and before 03/15/2021.
28.4\% of all messages (3,931,136) are forwarded messages, showing the popularity and relevance of this feature for Telegram. In addition to the 4,962 channels, we collected the metadata of 43,142 additional channels that were either the source of forwarded messages or were mentioned in a message.

39.2\% of all collected messages (5,421,845) are in German, which is the most frequent language, followed by English and Russian. 2,748 of the 4,962 crawled channels (55.4\%) are classified as German-speaking according to our approach. 

\subsection{Building Classification Models for Telegram Messages}

\paragraph{Models} Table \ref{tab:binary_classifiers_testset_results} presents the classification metrics of the six trained classification models. It comprises the precision, recall, and F1 score of the abusive class as well as the macro F1 score and the used model that performed best on the dataset.

\begin{table}[ht]
\centering
\caption{Classification performance of the classifiers}
\label{tab:binary_classifiers_testset_results}
\footnotesize
\begin{tabular}[ht]{lcccccc}
\hline
Dataset & Prec & Rec &  F1 & Macro F1 & Model\\
\hline
GermEval 18  & 71.1 & 61.0 & 65.7 & 75.0 & dbmdz\\
GermEval 19  & 72.2 & 85.1 & 78.1 &77.1 & dbmdz\\
GermEval 18/19  & 87.6 & 77.6 & 82.3 &  83.8 & dbmdz\\
HASOC 20 & 69.0 & 73.7 & 71.3 & 80.6 & deepset \\
HASOC 19/20  & 71.0 & 69.9 & 70.4 & 80.3 & dbmdz \\
COVID-19  & 73.9 & 69.9 & 71.8 & 82.3 & deepset \\
\hline
\end{tabular}
\end{table}

\begin{figure}[ht]
\centering
  \includegraphics[width=0.48\textwidth]{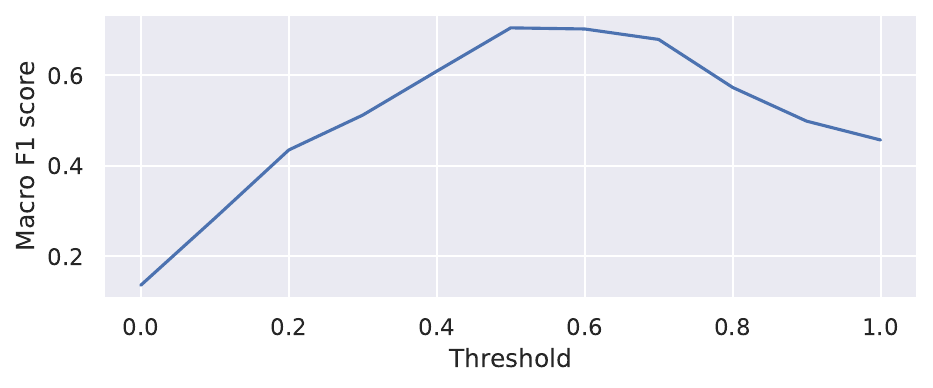}
  \caption{Macro F1 score dependent on various threshold for Perspective API on test set.}
  \label{fig:threshold_toxicity}
\end{figure}

\begin{figure*}[ht]
     \centering
     \begin{subfigure}[t]{0.42\textwidth}
         \centering
         \includegraphics[width=\textwidth]{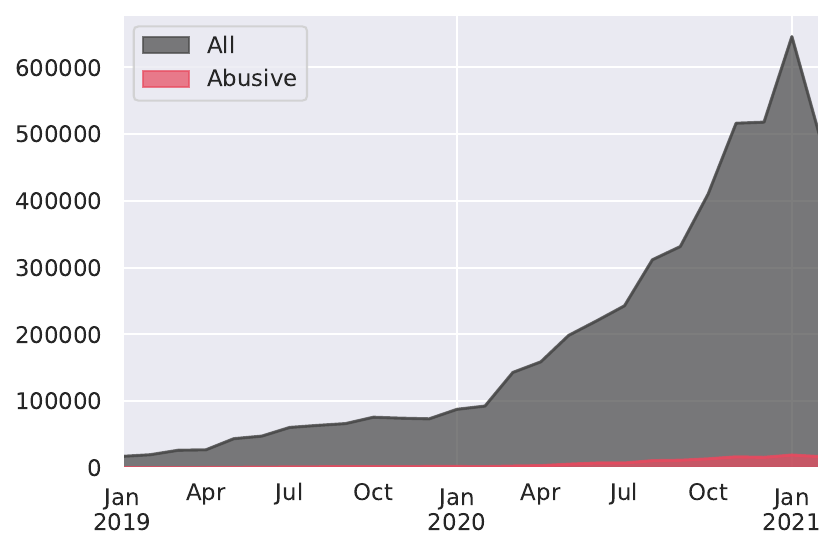}
         \caption{\textit{Absolute number of all and abusive messages from German channels.}}
         \label{fig:absolutenumber}
     \end{subfigure}
     \hfill
     \begin{subfigure}[t]{0.40\textwidth}
         \centering
         \includegraphics[width=\textwidth]{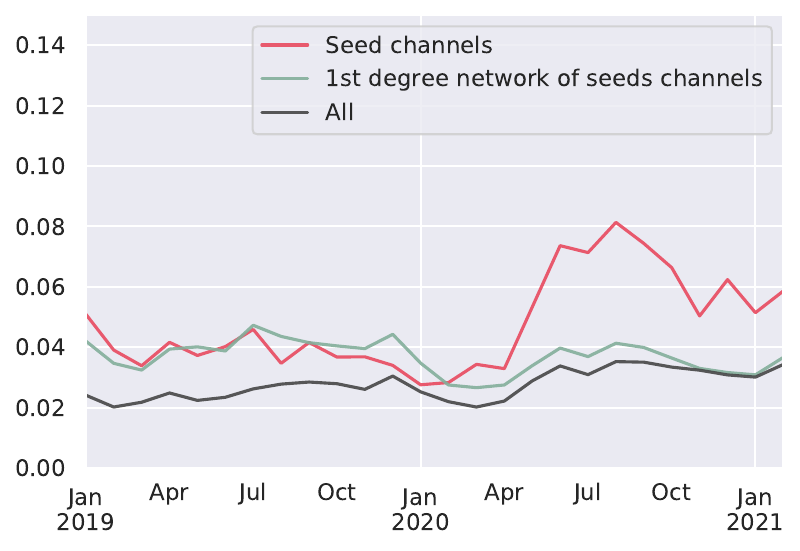}
         \caption{\textit{Relative share of abusive messages for German channels.}}
         \label{fig:relativenumber}
     \end{subfigure}
         \caption{Evolution of abusive messages in absolute and relative terms.}
        \label{fig:evolution}
\end{figure*}

\paragraph{Evaluating Classification Models} To test the trained classification models, we annotated 1,150 Telegram messages. One message was removed during the annotation process because it did not contain any text, resulting in 1,149 annotated messages. 968 (84.2\%) were labeled as \textit{neutral} and 126 (15.8\%) as \textit{abusive}. The Krippendorff's alpha was 73.87\%, which is a good inter-rater reliability score in the context of hate speech and abusive language \citep{kurrek-etal-2020-towards}. 

Figure \ref{fig:matrices} visualized the classification performance of the various classifiers on the evaluation set. It presents the confusion matrix, the F1 score of the abusive class, and the macro F1 score of the six trained classification models (a--f), the Perspective API (g), and the best combination of the six classifiers (h).
Let us first compare the six classification models that we trained on the different datasets.  The best-performing model is COVID-19; it outperformed the other models in terms of F1 score (54.95\%) and macro F1 score (71.91\%). In comparison to the COVID-19 test set, however, the performance drastically decreased. This should not be surprising because Telegram messages differ from tweets in terms of structure and content.

To benchmark the performance of our classification model, we used Google’s Perspective API to classify messages. The API returns a toxicity score between 0 and 1 for a text. We translated this value by setting a threshold. If the value is above or equal to the threshold, the label is \textit{abusive}; otherwise, the label is \textit{neutral}. We set the threshold to 0.5 because it produced the best macro F1 score. Figure \ref{fig:threshold_toxicity} shows the macro F1 scores dependent on various thresholds; the highest macro F1 score is achieved with a threshold of 0.5. Comparing the performance of the Perspective API with our best-performing model, our model has a higher F1 score  (54.95\% vs. 53.50\%) and macro F1 score (71.91\% vs 70.51\%). This is surprising because the Perspective API is built for comments more similar to Telegram messages than tweets.

Because the datasets cover different aspects of abusive language, we also examined whether a combination of all six classifiers can improve performance. Our finding is that a majority vote (at least four classifiers vote for \textit{abusive}) of all six models is the best-performing combination in terms of the macro F1 score, as shown in Figure \ref{fig:combination}. It outperforms the Perspective API and the classifier trained on the COVID-19 datatset in terms of macro F1 score. To validate the result, we applied the McNemar's test \cite{10.1162/089976698300017197} to show that the best combination performs significantly differently ($p<0.05$) from the Perspective API ($p=2.69\times 10^{-5}$) and COVID-19 ($p=1.02\times 10^{-3}$). Therefore, the best combination is the majority vote with at least four classifiers voting for \textit{abusive}, which we used for the following two case studies.


\paragraph{Analyzing the Evolution of Abusive Content}
Figure \ref{fig:relativenumber} shows how the number of messages in the German Telegram channels has increased between the beginning of 2019 and 2021. We can trace the growth of these channels back to the phenomenon of deplatforming. Deplatforming means that actors are permanently banned on traditional social media platforms (e.g., Facebook, Twitter, and YouTube), resulting in them moving to less or unregulated platforms (e.g., Telegram and Gab) \citep{roger-deplatforming,hatenotfound,urman-2020}. Notably, the increase in messages accelerated with the rise of COVID-19 (February 2020). The explanation is the same. Traditional social media platforms (e.g., Twitter and YouTube) blocked accounts of hate actors spreading conspiracy theories regarding COVID-19, causing migration to Telegram and alternative platforms \citep{hatenotfound,Holzer2021}. Simultaneously with the growing number of messages every month (black curve), abusive content also increased (red curve).

To answer the question of whether the abusive content has grown in the same manner, we plotted the relative share of abusive content in Figure \ref{fig:relativenumber}. The black line represents the relative share for all messages. We observe that the share of abusive content increased from 2.4\% to 3.4\% during the 26 months. The red line shows the portion of abusive messages in the seed channels. It is not surprising that the share is significantly higher because these channels were classified as hater channels by \citet{hatenotfound}. The line follows the trend: the abusive content of the selected channels is growing. The green line visualizing the percentage of abusive messages in the channels being in the first-degree network of the seed channels\footnote{A channel is in the first-degree network if a seed channel mentions the channel or forwards a message from this channel and vice versa.} does not show the trend. A potential explanation is that the number of channels in the first-degree network has increased over time, causing an alignment of the relative share with the overall average. Overall, the prevalence of abusive content for the entire period is 3.1\%, 5.3\% for the seed channels, and 3.5\% for the 1st degree network of the seed channels.

In summary, we observe the trend that messages classified as \textit{abusive} by our combined model increase in absolute and relative terms in the German hater community on Telegram.

\subsection{Building a Classification Model for the Hatefulness of Channels}
In this section, we report the results of our classification model for identifying hateful users, along with additional findings in the process of setting up our model.

\paragraph{Channel Labels} The dataset for developing a channel classification model contains 2,420 German channels that were active in 2020 and posted 3,232,721 messages. 809 of 2,420 channels (33.4\%) are labeled as \textit{hater}, the rest as \textit{neutral}. Each channel is represented by a node in the directed graph. In total, we identified 146,865 edges between channels. This leads to a density of 0.0251 and an average in- and out-degree of 60.73.

\begin{figure}[ht]
\centering
  \includegraphics[width=0.40\textwidth]{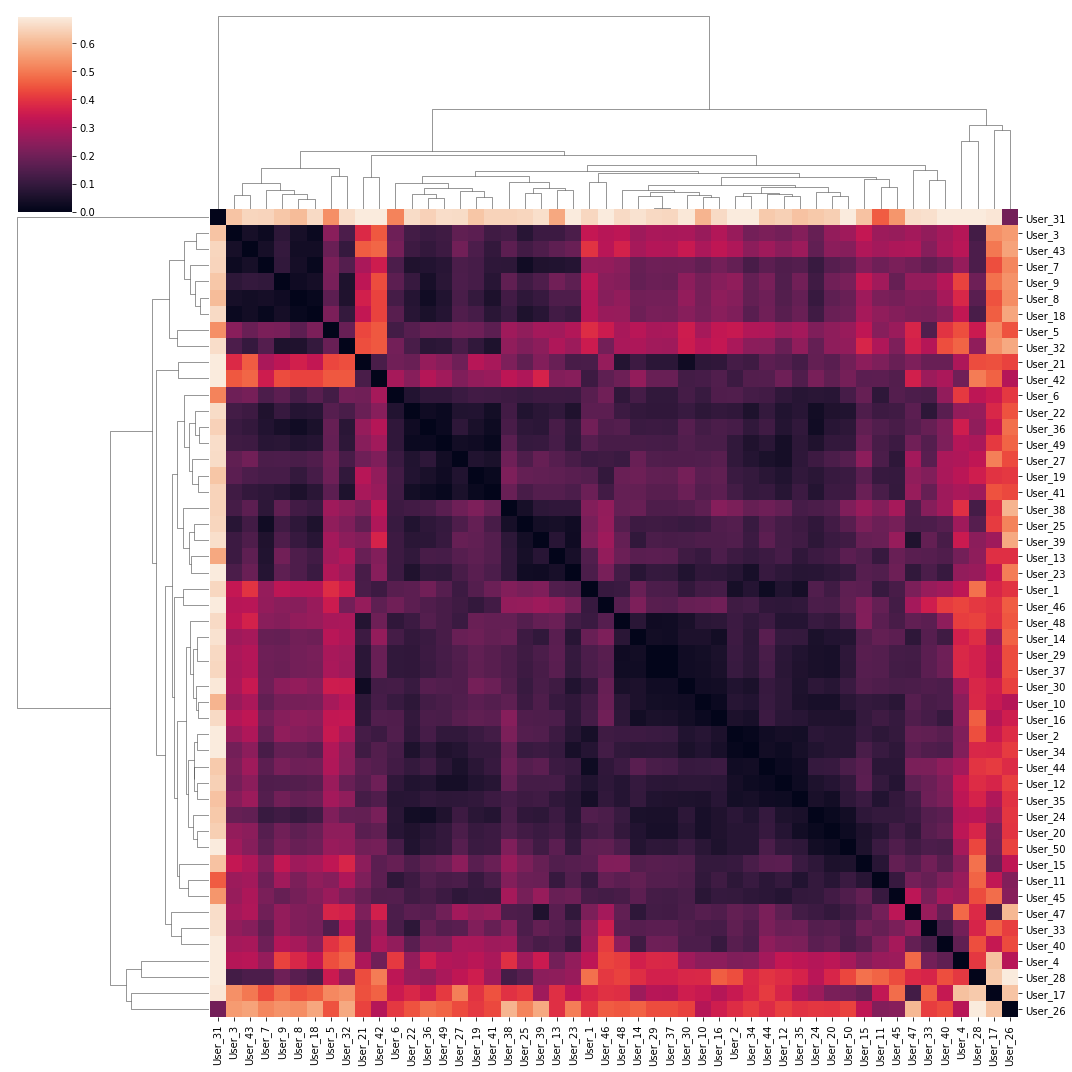}
  \caption{Similarity matrix for the seed channels of the Telegram dataset.}
  \label{fig:sim_matrix}
\end{figure}

\paragraph{Topical Distribution} As the first result, we examined clusters based on the topical distribution of the seed channels. To do this, the similarity between the topical distribution of each pair of users has been computed using the Jensen–-Shannon divergence. For the resulting similarity matrix, a hierarchical clustering approach has been used to group similar users into clusters, as described in Figure \ref{fig:sim_matrix}. While we only disclose an anonymized version of our results, we report that the upper left cluster consists only of sources for alternative news and the large cluster in the center mainly contains actors who belong to the far-right network.

\begin{figure*}[ht]
     \centering
     \begin{subfigure}[b]{0.42\textwidth}
         \centering
         \includegraphics[width=\textwidth]{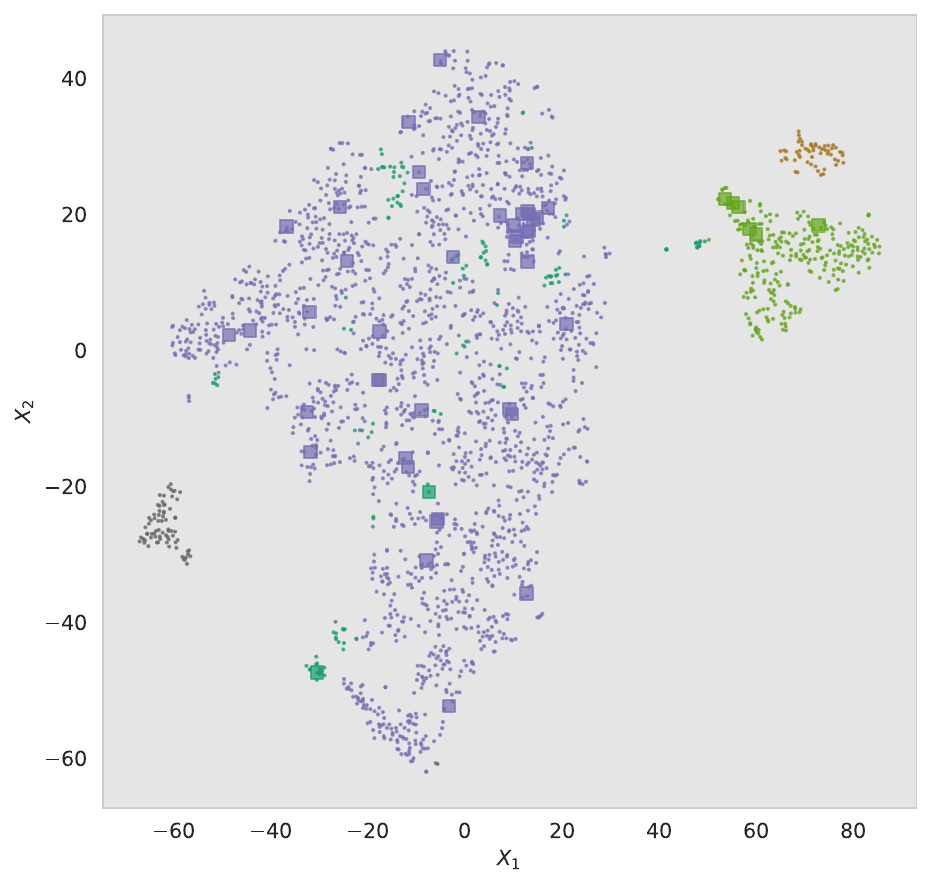}
         \caption{\textit{Graph embeddings with community labels}}
         \label{fig:emb_com}
     \end{subfigure}
     \hfill
     \begin{subfigure}[b]{0.42\textwidth}
         \centering
         \includegraphics[width=\textwidth]{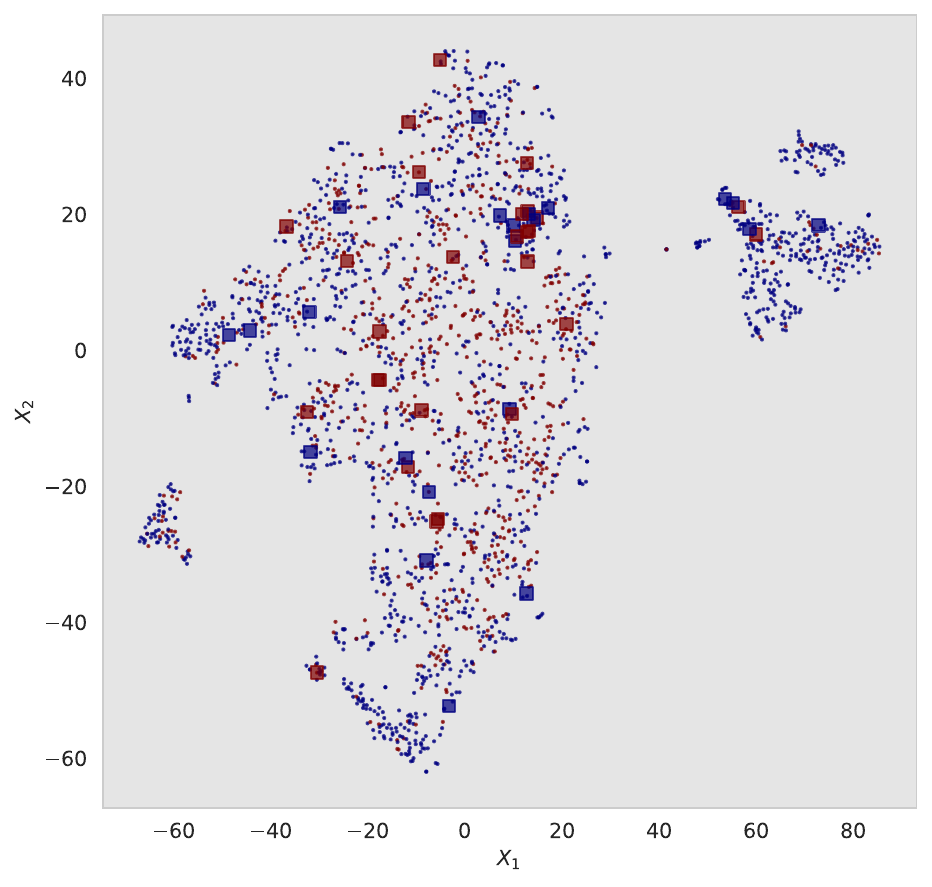}
         \caption{\textit{Graph embeddings with hate class labels (red=haters)}}
         \label{fig:emb_hate}
     \end{subfigure}
    \caption{Comparison of graph embeddings with community and hate class labels.}
    \label{fig:graph_embeddings}
\end{figure*}

\paragraph{Graph Embeddings} Before using the graph embeddings from the directed GraphSAGE model for the classification model, we investigate the expressiveness of the embeddings for community detection. For this, we applied the dimensional reduction method UMAP to our embeddings to find more dense representations. In the second step, we used DB-SCAN to cluster these reduced embeddings. In Figure \ref{fig:graph_embeddings}, we report the results of the community detection, along with visualization indicating the label of each node. Seed profiles are marked with a large square instead of a dot. The clustering algorithm recognizes four distinct communities along with one outlier class. The large community in the center does not only contain most of the seed channels in our dataset but also the largest proportion of channels labeled as \textit{hater} (38\%). In the other communities, we find a significantly lower proportion of hatefully classified users (5\%-24\%). In the outlier class, 33\% are hater. From that, we deduce that hateful users appear more often in communities with other hateful users.

\begin{figure}[ht]
\centering
  \includegraphics[width=0.35\textwidth]{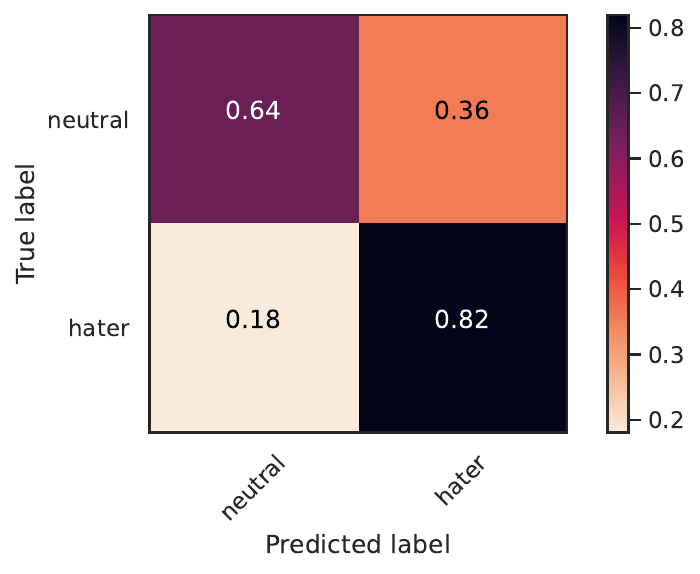}
  \caption{Confusion matrix of model to classify channels.}
  \label{fig:nodeclassification}
\end{figure}

\paragraph{Channel Classification} The classification model trained to distinguish between \textit{hater} and \textit{neutral} channels achieves a macro F1 score of 69.5\% (\textit{neutral}: 74.2\%; \textit{hater}: 64.9\%). Figure \ref{fig:nodeclassification} visualizes the confusion matrix of the classification model for the test set. We observe that the model performs well in predicting the labels of the German Telegram channels.   

\section{Discussion}

Regarding RQ1, we can state that existing abusive language datasets from Twitter can be used to develop an abusive language classification model for Telegram messages. However, we have to accept a decline in classification performance. Comparing the macro F1 scores of the classifiers on the original test and evaluation sets, we observe an average decline of approximately 12.5pp. To better assess this value, it is helpful to look into the study on the generalizability of abusive language datasets from \citet{swamy-etal-2019-studying}. They trained models on different abusive language datasets and evaluated them on each other. The average performance decline is 18.1pp if a classifier is evaluated on another test set. Considering this aspect, we can claim that our models perform decently, especially the combination of all six classification models with a threshold of four. This claim is supported by the fact that the combined models outperform the Perspective API in terms of F1 score. We integrated this external model provided by Google as a benchmark because it is developed to handle different types of texts (e.g., comments, posts, and emails), and it is in production \citep{perspective-case}. Consequently, we can state that our approach is successful, but it still provides room for improvement.

Regarding RQ2, we observe an increasing prevalence of abusive messages in the collected Telegram subnetwork, especially in the group of the seed channels. Notably, the rise of COVID-19 caused a significant increase. One may argue that the relative share of abusive content is unreliable because our combined classification model is imperfect. However, the change in the relative share provides a reliable indication of an increasing amount of abusive content. We trace this trend back to the deplatforming activities of large social media platforms and Telegram’s lack of moderation. We have to point out that the prevalence of abusive content is unrepresentative of the entire German Telegram network. Due to our snowball sampling approach, we have an obvious selection bias because we started with channels that were classified as hate actors by \citet{hatenotfound}. Nevertheless, we assume that the prevalence of abusive content is larger on Telegram than on traditional social media platforms, such as Twitter, Facebook, and YouTube, that have implemented reporting and monitoring processes. In the case of Telegram, such processes are missing.

Regarding RQ3, we developed a classification model to predict whether a channel is a hate actor. It uses the network structure and the topic distribution of messages in each channel for prediction. Our model achieves a macro F1 score of 69.5\%. To the best of our knowledge, we are the first to develop such a classification model for Telegram channels. Therefore, we do not have a baseline to compare our results with. However, \citet{Ribeiro2018} and \citet{liuneighbours} developed comparable models classifying Twitter accounts as hateful or normal. For the same dataset, \citet{Ribeiro2018} and \citet{liuneighbours} achieved F1 scores of 67.0\% and 79.9\%, respectively, for hateful users. Our F1 score of 64.9\% is not directly comparable with these results, but it is in a similar order of magnitude, supporting our approach.

Addressing RQ4, we presented two approaches that allow clustering channels: The first approach leverages the topical distribution of channels to group actors based on the topical similarity of the content they spread. Applying this to the seed channels for the collection of the dataset indicates promising results for future research attempts in clustering actors on social media based on the content of their postings in a time-saving manner. The second method we propose in this context leverages embeddings learned from the social graph that we developed from the dataset. It also uses the results from the topic model, i.e., it merges relational data between the channels with the content they shared. Our results indicate different communities that vary in the number of hateful users. Large communities appear to be spanned by seed users;  however, we also detected smaller communities that do not contain any seed users, indicating that our sampling approach could find new user clusters. For a more precise evaluation of these results, more general information about the German hater community would have been helpful.

\section{Conclusion and Future Work}

To the best of our knowledge, we are the first to develop abusive language classification models for German messages on Telegram. Our results look promising. The text model outperforms Google's Perspective API in terms of F1 score (macro F1: 73.2\%). Similarly, the channel classification model provides good performance in detecting \textit{hater} channels (macro F1: 69.5\%). In addition, we have outlined methods for facilitating and scaling abusive language analysis on a message level as well as on a channel level. In the latter case, we fully relied on unsupervised learning methods, which makes these approaches particularly appealing. Furthermore, we published the first abusive language dataset consisting of German Telegram messages. However, we see room for improvement and potential for future work.

The research community would benefit from larger annotated corpora, including media files shared in Telegram channels (e.g., photos with messages, memes, and videos). Because such media files (e.g., memes) are used to transport hate \cite{kiela2021hateful}, they are relevant for the problem of detecting abusive content but were not part of this study.

Regarding the classification model for \textit{hater} channels, integrating additional data (e.g., metadata of the channels) and enhancing the NN architecture could improve classification performance. An explorative network analysis of the subnetwork could help identify additional features. In addition, a larger portion of Telegram should be collected with other seed users to mitigate the selection bias introduced by our hateful seed users.

We also encourage researchers from various core disciplines, such as machine learning and social sciences, to synergize to validate the performances achieved by sophisticated learning frameworks applied to large amounts of data. Due to the unstoppable increase in content produced on social platforms such as Telegram, automatic methods for generating insights will become indispensable. Finally, the hate speech detection community should look into applying approaches such as ours to alternative social media platforms as hate actors will congregate there as deplatforming efforts continue.

\bibliography{ref}

\begin{thebibliography}{38}
\providecommand{\natexlab}[1]{#1}
\providecommand{\url}[1]{\texttt{#1}}
\providecommand{\urlprefix}{URL }
\expandafter\ifx\csname urlstyle\endcsname\relax
  \providecommand{\doi}[1]{doi:\discretionary{}{}{}#1}\else
  \providecommand{\doi}{doi:\discretionary{}{}{}\begingroup
  \urlstyle{rm}\Url}\fi

\bibitem[{Angelov(2020)}]{angelov2020top2vec}
Angelov, D. 2020.
\newblock {Top2vec: Distributed representations of topics}.
\newblock \emph{arXiv preprint arXiv:2008.09470} .

\bibitem[{Baumgartner et~al.(2020)Baumgartner, Zannettou, Squire, and
  Blackburn}]{baumgartner2020pushshift}
Baumgartner, J.; Zannettou, S.; Squire, M.; and Blackburn, J. 2020.
\newblock {The Pushshift Telegram Dataset}.
\newblock In \emph{Proceedings of the International AAAI Conference on Web and
  Social Media}, volume~14, 840--847.

\bibitem[{Bretschneider and Peters(2017)}]{bretschneider2017detecting}
Bretschneider, U.; and Peters, R. 2017.
\newblock {Detecting offensive statements towards foreigners in social media}.
\newblock In \emph{Proceedings of the 50th Hawaii International Conference on
  System Sciences}.

\bibitem[{Chan, Schweter, and M{\"o}ller(2020)}]{chan-etal-2020-germans}
Chan, B.; Schweter, S.; and M{\"o}ller, T. 2020.
\newblock {German's Next Language Model}.
\newblock In \emph{Proceedings of the 28th International Conference on
  Computational Linguistics}, 6788--6796. Barcelona, Spain (Online):
  International Committee on Computational Linguistics.

\bibitem[{{CSIRO's Data61}(2018)}]{StellarGraph}
{CSIRO's Data61}. 2018.
\newblock {StellarGraph Machine Learning Library}.
\newblock \url{https://github.com/stellargraph/stellargraph}.

\bibitem[{Devlin et~al.(2019)Devlin, Chang, Lee, and
  Toutanova}]{devlin-etal-2019-bert}
Devlin, J.; Chang, M.-W.; Lee, K.; and Toutanova, K. 2019.
\newblock {BERT: Pre-training of Deep Bidirectional Transformers for Language
  Understanding}.
\newblock In \emph{Proceedings of the 2019 Conference of the North {A}merican
  Chapter of the Association for Computational Linguistics: Human Language
  Technologies, Volume 1 (Long and Short Papers)}, 4171--4186. Minneapolis,
  Minnesota: Association for Computational Linguistics.

\bibitem[{Dietterich(1998)}]{10.1162/089976698300017197}
Dietterich, T.~G. 1998.
\newblock {Approximate Statistical Tests for Comparing Supervised
  Classification Learning Algorithms}.
\newblock \emph{Neural Computation} 10(7): 1895--1923.
\newblock ISSN 0899-7667.
\newblock \doi{10.1162/089976698300017197}.
\newblock \urlprefix\url{https://doi.org/10.1162/089976698300017197}.

\bibitem[{Duggan(2017)}]{duggan2017online}
Duggan, M. 2017.
\newblock \emph{{Online harassment 2017}}.
\newblock Pew Research Center.

\bibitem[{Echikson and Knodt(2018)}]{echikson2018germany}
Echikson, W.; and Knodt, O. 2018.
\newblock {Germany’s NetzDG: A key test for combatting online hate}.
\newblock \emph{CEPS Policy Insight} .

\bibitem[{Eckert, Leipertz, and Schmidt(2021)}]{Eckert2021}
Eckert, S.; Leipertz, S.; and Schmidt, C. 2021.
\newblock {Querdenker: Wie die Corona-Krise zu Radikalisierung f\"uhrte}.
\newblock \emph{Norddeutscher Rundfunk}
  \urlprefix\url{https://story.ndr.de/querdenker/}.
\newblock Visited on 11/20/2021.

\bibitem[{Fielitz and Schwarz(2020)}]{hatenotfound}
Fielitz, M.; and Schwarz, K. 2020.
\newblock \emph{{Hate not Found?! Deplatforming the Far-Right and its
  Consequences}}.
\newblock Institut f\"ur Demokratie und Zivilgesellschaft: Jena.

\bibitem[{Grave et~al.(2018)Grave, Bojanowski, Gupta, Joulin, and
  Mikolov}]{grave2018learning}
Grave, E.; Bojanowski, P.; Gupta, P.; Joulin, A.; and Mikolov, T. 2018.
\newblock {Learning Word Vectors for 157 Languages}.
\newblock In \emph{Proceedings of the International Conference on Language
  Resources and Evaluation (LREC 2018)}.

\bibitem[{Hamilton, Ying, and Leskovec(2017)}]{hamilton2017inductive}
Hamilton, W.~L.; Ying, R.; and Leskovec, J. 2017.
\newblock {Inductive representation learning on large graphs}.
\newblock In \emph{Proceedings of the 31st International Conference on Neural
  Information Processing Systems}, 1025--1035.

\bibitem[{Hohlfeld et~al.(2021)Hohlfeld, Bauerfeind, Braglia, Butt, Dietz,
  Drexel, Fedlmeier, Fischer, Gandl, Glaser, Haberzettel, Helling, Käsbauer,
  Kast, Krieger, Lächner, Malkanova, Raab, Rech, and Weymar}]{Hohlfeld2021}
Hohlfeld, R.; Bauerfeind, F.; Braglia, I.; Butt, A.; Dietz, A.-L.; Drexel, D.;
  Fedlmeier, J.; Fischer, L.; Gandl, V.; Glaser, F.; Haberzettel, E.; Helling,
  T.; Käsbauer, I.; Kast, M.; Krieger, A.; Lächner, A.; Malkanova, A.; Raab,
  M.-K.; Rech, A.; and Weymar, P. 2021.
\newblock {Communicating COVID-19 against the backdrop of conspiracy
  ideologies: How public figures discuss the matter on Facebook and Telegram} .

\bibitem[{Holzer(2021)}]{Holzer2021}
Holzer, B. 2021.
\newblock {Zwischen Protest und Parodie : Strukturen der
  Querdenken-Kommunikation auf Telegram (und anderswo)}.
\newblock In Reichardt, S., ed., \emph{{Die Misstrauensgemeinschaft der
  Querdenker : Die Corona-Proteste aus kultur- und sozialwissenschaftlicher
  Perspektive}}, 125--157. Frankfurt: Campus Verlag.

\bibitem[{Honnibal et~al.(2020)Honnibal, Montani, Van~Landeghem, and
  Boyd}]{spacy}
Honnibal, M.; Montani, I.; Van~Landeghem, S.; and Boyd, A. 2020.
\newblock {spaCy: Industrial-strength Natural Language Processing in Python}.
\newblock \urlprefix\url{https://doi.org/10.5281/zenodo.1212303}.

\bibitem[{Jigsaw(2021)}]{perspective-case}
Jigsaw. 2021.
\newblock {Perspective API - Case Studies}
  \urlprefix\url{https://www.perspectiveapi.com/case-studies/}.
\newblock Visited on 11/20/2021.

\bibitem[{Kiela et~al.(2021)Kiela, Firooz, Mohan, Goswami, Singh, Fitzpatrick,
  Bull, Lipstein, Nelli, Zhu et~al.}]{kiela2021hateful}
Kiela, D.; Firooz, H.; Mohan, A.; Goswami, V.; Singh, A.; Fitzpatrick, C.~A.;
  Bull, P.; Lipstein, G.; Nelli, T.; Zhu, R.; et~al. 2021.
\newblock {The Hateful Memes Challenge: Competition Report}.
\newblock In \emph{NeurIPS 2020 Competition and Demonstration Track}, 344--360.
  PMLR.

\bibitem[{{Kili Technology}(2021)}]{kili}
{Kili Technology}. 2021.
\newblock {Text annotation tool}.
\newblock \urlprefix\url{https://kili-technology.com}.
\newblock Visited on 11/20/2021.

\bibitem[{Krippendorff(2004)}]{krippendorff2004content}
Krippendorff, K. 2004.
\newblock \emph{{Content Analysis: An Introduction to Its Methodology}}.
\newblock Content Analysis: An Introduction to Its Methodology. Sage.

\bibitem[{Kurrek, Saleem, and Ruths(2020)}]{kurrek-etal-2020-towards}
Kurrek, J.; Saleem, H.~M.; and Ruths, D. 2020.
\newblock {Towards a Comprehensive Taxonomy and Large-Scale Annotated Corpus
  for Online Slur Usage}.
\newblock In \emph{Proceedings of the Fourth Workshop on Online Abuse and
  Harms}, 138--149. Online: Association for Computational Linguistics.

\bibitem[{Li et~al.(2021)Li, Zaidi, Liu, and Li}]{liuneighbours}
Li, S.; Zaidi, N.~A.; Liu, Q.; and Li, G. 2021.
\newblock {Neighbours and Kinsmen: Hateful Users Detection with Graph Neural
  Network}.
\newblock In Karlapalem, K.; Cheng, H.; Ramakrishnan, N.; Agrawal, R.~K.;
  Reddy, P.~K.; Srivastava, J.; and Chakraborty, T., eds., \emph{Advances in
  Knowledge Discovery and Data Mining}, 434--446. Cham: Springer International
  Publishing.

\bibitem[{Mandl et~al.(2020)Mandl, Modha, Kumar~M, and
  Chakravarthi}]{hasoc2020}
Mandl, T.; Modha, S.; Kumar~M, A.; and Chakravarthi, B.~R. 2020.
\newblock {Overview of the HASOC Track at FIRE 2020: Hate Speech and Offensive
  Language Identification in Tamil, Malayalam, Hindi, English and German}.
\newblock In \emph{Forum for Information Retrieval Evaluation}, FIRE 2020,
  29--32. New York, NY, USA: Association for Computing Machinery.

\bibitem[{Mandl et~al.(2019)Mandl, Modha, Majumder, Patel, Dave, Mandlia, and
  Patel}]{hasoc2019}
Mandl, T.; Modha, S.; Majumder, P.; Patel, D.; Dave, M.; Mandlia, C.; and
  Patel, A. 2019.
\newblock {Overview of the HASOC Track at FIRE 2019: Hate Speech and Offensive
  Content Identification in Indo-European Languages}.
\newblock In \emph{Proceedings of the 11th Forum for Information Retrieval
  Evaluation}, FIRE '19, 14–17. New York, NY, USA: Association for Computing
  Machinery.
\newblock ISBN 9781450377508.

\bibitem[{M{\"u}ller and Schwarz(2021)}]{muller2021fanning}
M{\"u}ller, K.; and Schwarz, C. 2021.
\newblock {Fanning the flames of hate: Social media and hate crime}.
\newblock \emph{Journal of the European Economic Association} 19(4):
  2131--2167.

\bibitem[{Rafael(2019)}]{rafael2019}
Rafael, Simone;~Ritzmann, A. 2019.
\newblock \emph{{Hate Speech and Radicalisation Online - The OCCI Research
  Report}}, chapter Background: the ABC of hate speech, extremism and the
  NetzDG.
\newblock ISD Global.

\bibitem[{{R{\"a}ther}(2021)}]{svenja2021}
{R{\"a}ther}, S. 2021.
\newblock \emph{{Investigating Techniques for Learning with Limited Labeled
  Data for Hate Speech Classification}}.
\newblock Master's thesis, Technical University of Munich.
\newblock Advised and supervised by Maximilian Wich and Georg Groh.

\bibitem[{Ribeiro et~al.(2018)Ribeiro, Calais, Santos, Almeida, and
  Meira~Jr}]{Ribeiro2018}
Ribeiro, M.; Calais, P.; Santos, Y.; Almeida, V.; and Meira~Jr, W. 2018.
\newblock {Characterizing and Detecting Hateful Users on Twitter }.
\newblock In \emph{Proceedings of the Twelfth International AAAI Conference on
  Web and Social Media (ICWSM 2018)}.

\bibitem[{Rogers(2020)}]{roger-deplatforming}
Rogers, R. 2020.
\newblock {Deplatforming: Following extreme Internet celebrities to Telegram
  and alternative social media}.
\newblock \emph{European Journal of Communication} 35(3): 213--229.

\bibitem[{Ross et~al.(2016)Ross, Rist, Carbonell, Cabrera, Kurowsky, and
  Wojatzki}]{ross2016hatespeech}
Ross, B.; Rist, M.; Carbonell, G.; Cabrera, B.; Kurowsky, N.; and Wojatzki, M.
  2016.
\newblock {Measuring the Reliability of Hate Speech Annotations: The Case of
  the European Refugee Crisis}.
\newblock In Bei{\ss}wenger, M.; Wojatzki, M.; and Zesch, T., eds.,
  \emph{Proceedings of NLP4CMC III: 3rd Workshop on Natural Language Processing
  for Computer-Mediated Communication}, volume~17 of \emph{Bochumer
  Linguistische Arbeitsberichte}, 6--9. Bochum.

\bibitem[{Solopova, Scheffler, and Popa-Wyatt(2021)}]{solopova2021telegram}
Solopova, V.; Scheffler, T.; and Popa-Wyatt, M. 2021.
\newblock {A Telegram corpus for hate speech, offensive language, and online
  harm}.
\newblock \emph{Journal of Open Humanities Data} 7.

\bibitem[{{Stru{\ss}} et~al.(2019){Stru{\ss}}, {Siegel}, {Ruppenhofer},
  {Wiegand}, and {Klenner}}]{struss2019overview}
{Stru{\ss}}, J.~M.; {Siegel}, M.; {Ruppenhofer}, J.; {Wiegand}, M.; and
  {Klenner}, M. 2019.
\newblock {Overview of GermEval Task 2, 2019 shared task on the identification
  of offensive language}.
\newblock In \emph{{Proceedings of the 15th Conference on Natural Language
  Processing (KONVENS 2019)}}, 354--365.

\bibitem[{Swamy, Jamatia, and Gamb{\"a}ck(2019)}]{swamy-etal-2019-studying}
Swamy, S.~D.; Jamatia, A.; and Gamb{\"a}ck, B. 2019.
\newblock {Studying Generalisability across Abusive Language Detection
  Datasets}.
\newblock In \emph{Proceedings of the 23rd Conference on Computational Natural
  Language Learning (CoNLL)}, 940--950. Hong Kong, China: Association for
  Computational Linguistics.

\bibitem[{Urman and Katz(2020)}]{urman-2020}
Urman, A.; and Katz, S. 2020.
\newblock {What they do in the shadows: examining the far-right networks on
  Telegram}.
\newblock \emph{Information, Communication \& Society} 0(0): 1--20.

\bibitem[{Wich et~al.(2021)Wich, Breitinger, Strathern, Naimarevic, Groh, and
  Pfeffer}]{wich2021}
Wich, M.; Breitinger, M.; Strathern, W.; Naimarevic, M.; Groh, G.; and Pfeffer,
  J. 2021.
\newblock {Are your Friends also Haters? Identification of Hater Networks on
  Social Media: Data Paper}.
\newblock In \emph{Companion Proceedings of the Web Conference 2021 (WWW'21
  Companion)}.

\bibitem[{Wich, R\"ather, and Groh(2021)}]{wich-covid-19}
Wich, M.; R\"ather, S.; and Groh, G. 2021.
\newblock {German Abusive Language Dataset with Focus on COVID-19}.
\newblock In \emph{{Proceedings of the 17th Conference on Natural Language
  Processing (KONVENS 2021)}}.

\bibitem[{Wiegand, Siegel, and Ruppenhofer(2018)}]{wiegand2018overview}
Wiegand, M.; Siegel, M.; and Ruppenhofer, J. 2018.
\newblock {Overview of the germeval 2018 shared task on the identification of
  offensive language}.
\newblock In \emph{{Proceedings of the 14th Conference on Natural Language
  Processing (KONVENS 2018)}}.

\bibitem[{Williams et~al.(2020)Williams, Burnap, Javed, Liu, and
  Ozalp}]{williams2020hate}
Williams, M.~L.; Burnap, P.; Javed, A.; Liu, H.; and Ozalp, S. 2020.
\newblock {Hate in the Machine: Anti-Black and Anti-Muslim Social Media Posts
  as Predictors of Offline Racially and Religiously Aggravated Crime}.
\newblock \emph{The British Journal of Criminology} 60(1): 93--117.

\end{thebibliography}
\end{document}